# Morphological Reconstruction for Word Level Script Identification

**B.V.Dhandra**　　　　　　　　　　　　　　　　　　dhandra_b_v@yahoo.co.in
*P.G.Department of Studies and Research in*
*Computer Science, Gulbarga University,*
*Gulbarga -585106, India*

**Mallikarjun Hangarge**　　　　　　　　　　　　　　mhangarge@yahoo.co.in
*P.G.Department of Studies and Research in*
*Computer Science, Gulbarga University,*
*Gulbarga -585106, India*

**Abstract**

A line of a bilingual document page may contain text words in regional language and numerals in English. For Optical Character Recognition (OCR) of such a document page, it is necessary to identify different script forms before running an individual OCR system. In this paper, we have identified a tool of morphological opening by reconstruction of an image in different directions and regional descriptors for script identification at word level, based on the observation that every text has a distinct visual appearance. The proposed system is developed for three Indian major bilingual documents, Kannada, Telugu and Devnagari containing English numerals. The nearest neighbour and k-nearest neighbour algorithms are applied to classify new word images. The proposed algorithm is tested on 2625 words with various font styles and sizes. The results obtained are quite encouraging

**Keywords:** Script identification, Bilingual documents, OCR, Morphological reconstruction, regional descriptors

## 1. INTRODUCTION

As the world moves closer to the concept of the "paperless office," more and more communication and storage of documents is performed digitally. Documents and files that were once stored physically on paper are now being converted into electronic form in order to facilitate quicker additions, searches, and modifications, as well as to prolong the life of such records. A great portion of business documents and communication, however, still takes place in physical form and the fax machine remains a vital tool of communication worldwide. Because of this, there is a great demand for software, which automatically extracts, analyzes, and stores information from physical documents for later retrieval. All of these tasks fall under the general heading of document analysis, which has been a fast growing area of research in recent years.

A very important area in the field of document analysis is that of optical character recognition (OCR), which is broadly defined as the process of recognizing either printed or handwritten text from document images and converting it into electronic form. To date, many algorithms have been presented in the literature to perform this task for a specific language, and these OCRs will not work for a document containing more than one language/script.

Due to the diversity of languages and scripts, English has proved to be the binding language for India. Therefore, a bilingual document page may contain text words in regional language and numerals in English. So, bilingual OCR is needed to read these documents. To make a bilingual OCR successful, it is necessary to separate portions of different script regions of the bilingual document at word level and then identify the different script forms before running an individual OCR system. Among the works reported in this direction to distinguish between various Indian languages/scripts at word level are due to [4, 9, 12 and 14]. The algorithms proposed by Dhanya et al. [4] based on Gabor filters and spatial spread features, Padma et al. [9] based on discriminating features and Pal et al. [12] based on water reservoir and conventional features have recognition rate of more than 95%. The recognition accuracy of these algorithms falls drastically when the word length is less than three characters. Hence, the algorithms are word size dependent. Peeta Basa Pati et al. [14] have proposed word level script identification for Tamil, Devnagari and Oriya scripts based on 32 features using Gabor filters. It is obvious that the time complexity of this algorithm is more as it depends on 32 features. Authors have not reported about the performance of their algorithm for various font sizes and styles. Furthermore, the algorithms discussed so far have addressed only alphabet-based script identification (i.e. English text words separation from bilingual documents), whereas numeral script identification (i.e. English numerals separation from bilingual documents) is ignored. But, the fact is that, the large number of bilingual documents contains text words in regional languages and numerals in English (printed or handwritten). For example, Newspapers, Magazines, Books, Application forms, Railway Reservation forms etc. This has motivated us to develop a method for automatic script identification by separating English numerals (printed and handwritten) from bilingual documents.

From the literature survey, it is evident that some amount of work other than word level script/language identification has been carried out. Spitz [16] proposed a method for distinguishing between Asian and European languages by examining the upward concavities of connected components. Tan et al. [10] proposed a method based on texture analysis for automatic script and language identification from document images using multiple channel (Gabor) filters and Gray level co-occurrence matrices for seven languages: Chinese, English, Greek, Koreans, Malayalam, Persian and Russian. Hochberg, et al. [5] described a method of automatic script identification from document images using cluster-based templates. Tan [18] developed rotation invariant features extraction method for automatic script identification for six languages. Wood et al. [19] described projection profile method to determine Roman, Russian, Arabic, Korean and Chinese characters. Chaudhuri et al. [1] discussed an OCR system to read two Indian languages scripts: Bangla and Devnagari (Hindi). Chaudhuri et al. [2] described a complete printed Bangla OCR. Pal et al. [11] proposed an automatic technique of separating the text lines from 12 Indian scripts. Gaurav et al. [3] proposed a method for identification of Indian languages by combining Gabor filter based techniques and direction distance histogram classifier for Hindi, English, Malayalam, Bengali, Telugu and Urdu. Basavaraj et al. [13] proposed a neural network based system for script identification of Kannada, Hindi and English. Nagabhushan et al. [15] discussed an intelligent pin code script identification methodology based on texture analysis using modified invariant moments. In this paper an attempt is made to demonstrate the potentiality of morphological reconstruction approach for script identification at word level.

In Section 2, the brief overview of data collection, pre-processing and line and word segmentation are presented. In Section 3, the feature extraction, features computation and K nearest neighbour classifier are discussed. In Section 4, the proposed algorithm is presented. The experimental details and results obtained are presented in Section 5. Conclusion is given in Section 6.

B.V.Dhandra and Mallikarjun Hangarge

## 2. DATA COLLECTION AND PREPROCESSING

In this paper three data sets are used for experimentation. The first data set is of 500 document pages of Kannada, Telugu, Devnagari are obtained from various magazines, newspapers, books and other such documents containing variable font styles and sizes. The second data set is of 175 handwritten English numerals collected from 100 writers. The third data set consists of 150 word images of varied font styles and sizes. The collected documents are scanned using HP Scanner at 300 DPI, which usually yields a low noise and good quality document image. The digitized images are in gray tone and we have used Otsu's [22] global thresholding approach to convert them into two-tone images. The two-tone images are then converted into 0-1 labels where the label 1 represents the object and 0 represents the background. The small objects like, single or double quotation marks, hyphens and periods etc. are removed using morphological opening. The next step in pre-processing is skew detection and correction. Using algorithm [7], with minor modification (i.e. vertical dilation with line structuring element of length 10 pixels) skew detection and correction has been performed before segmentation.

### 2.1 Line and word segmentation

To segment the document image into several text lines, we use the valleys of the horizontal projection computed by a row-wise sum of black pixels. The position between two consecutive horizontal projections where the histogram height is least denotes one boundary line. Using these boundary lines, document image is segmented into several text lines. Similarly, to segment each text line into several text words, we use the valleys of the vertical projection of each text line obtained by computing the column-wise sum of black pixels. The position between two consecutive vertical projections where the histogram height is least denotes one boundary line. Using these boundary lines, every text line is segmented into several text words. The word wise segmentation is illustrated in Fig.1. These word images are then used to compute the eight-connected components of white pixels on the image and produce the bounding box for each of the connected components.

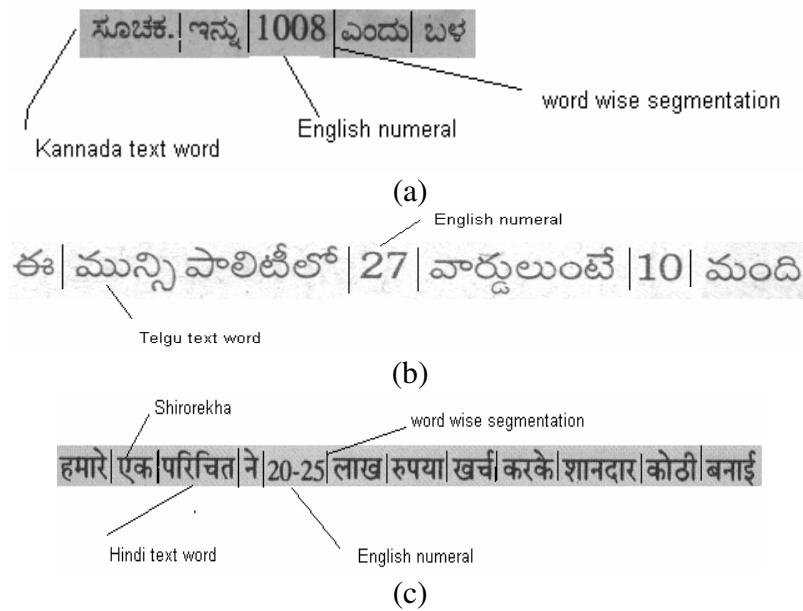

(a)

(b)

(c)

**FIGURE 1:** Word-Wise Segmentation of (a) Kannada (b) Telugu and (c) Devnagari Scripts

## 3. FEATURE EXTRACTION

Each sample or pattern that we attempt to classify is a word. It is helpful to study the general characteristics of each of the three proposed scripts for feature extraction.

**Devnagari:** Most of the characters of Devnagari script have a horizontal line at the upper part. In Devnagari, this line is called sirorekha. However, we shall call them as headlines. When two or more Devnagari characters sit side by side to form a word, the sirorekha or headline will touch each other and generates a big headline [10].

**Roman numerals (English numerals):** The important property of the Roman numerals (English) is the existence of the vertical strokes in its characters and has less number of horizontal strokes. By the experiment, it is noticed that the vertical strokes in digits like 1, 3, 4, 6, 8, 9, and 0 are more dominant than that of horizontal strokes as compared to Devnagari script.

**Kannada and Telugu Scripts**: The Kannada and Telugu scripts have more horizontal, right and left diagonal strokes. Thus, the right and left diagonal strokes will also play an important rule in distinguishing Kannada, Telugu, and Devnagari script from Roman numerals. These directional stroke features are extracted from the connected components of an image or pattern using morphological opening. In the following, we describe the features and their method of computation. To extract the characters or components containing strokes in vertical, horizontal, right and left diagonal directions, we have performed the erosion operation on the input binary image with the line-structuring element. The length of the structuring element is thresholded to 70 %( experimentally fixed) of average height of all the connected components of an image. The resulting image is used for morphological opening in four directions to obtain the strokes of an input image, as illustrated in Fig.2.

**Morphological Reconstruction:** Reconstruction is a morphological transformation involving two images and a structuring element. One image, the marker, is the starting point for the transformation. The other mask image constraints the transformation. In this paper, a fast hybrid reconstruction algorithm [8] is used for reconstruction and erode image is used as the marker image throughout the experiment.

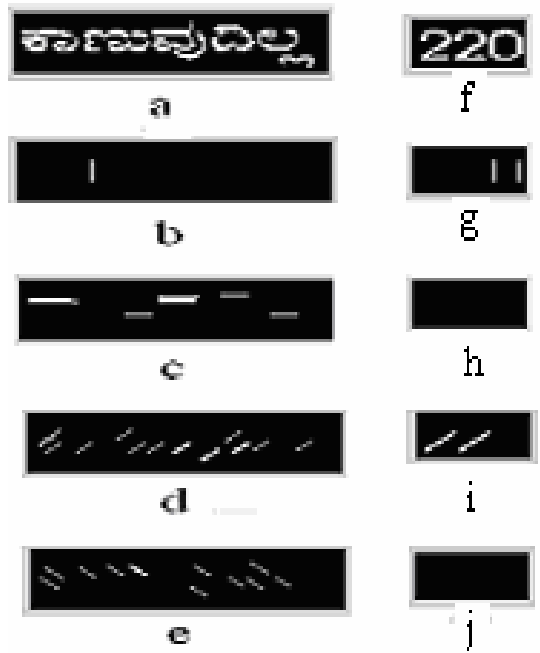

**FIGURE 2:** Shows the Stroke Extraction Process (a) Input Image of Kannada Script and (f) English Numeral (b), (c), (d), (e) and (g), (h), (i), (j) are Vertical, Horizontal, Right and Left Diagonal Stokes of Kannada and English Scripts Respectively

**Opening by Reconstruction:** It restores the shapes of the objects that remain after erosion. As an illustration the opening by reconstruction of Kannada word in vertical and horizontal directions is shown in Figure 3.

B.V.Dhandra and Mallikarjun Hangarge

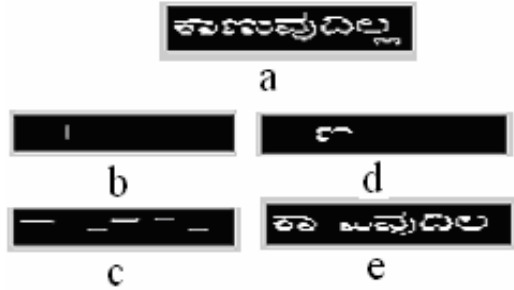

**FIGURE 3**: (a) Kannada Script Input Word Image, (b**)** Vertical Opening of (a), (c) Horizontal Opening of (a), (d) Vertical Reconstruction based on (b), (e) Horizontal Reconstruction based on (c).

For fill holes, we choose the marker image (erode image), $f_m$, to be 0 every where except on the image border, where it is set to 1-f. Here f is the image of a connected component.

$$f_m(x, y) = \begin{cases} 1 - f(x, y), & \text{if } (x, y) \text{ is on the border of } f \\ 0, & \text{otherwise} \end{cases}$$

Then $g = [R_f^c (f_m)]^c$ has the effect of the filling the holes in f, where, $R_f^c$ is the reconstructed image of f. As an example the horizontal reconstruction with fill holes of the Kannada script is shown in Figure 4.

$g = $ 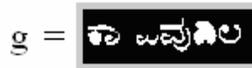

**FIGURE 4**: Represents Horizontal Reconstruction with Fill Holes**.**

**3.1 Features Computation**
1 On pixels densities (OPD) after reconstruction in vertical, horizontal, right and left diagonal directions with fill holes is given by

$$OPD(\theta) = \frac{\sum onpixels(g)}{size(g)}$$

where, θ varies from 0 to 135 with the incremental bandwidth of 45 degrees. The OPD values are real numbers. Thus, we obtain a set of four features by reconstruction approach. The remaining four features considered are aspect ratio, pixel ratio, eccentricity and extent. These features computation is discussed in the following. Throughout, the discussion of section 3.1, N is referred for the number of components present in an image.

2 Aspect Ratio: - The ratio of the height to the width of a connected component of an image [6]. The average aspect ratio (AAR) is defined as

$$AAR(pattern) = \frac{1}{N} \sum_{i}^{N} \frac{height(component_i)}{width(component_i)}$$

The value of AAR is a real number. Note that the aspect ratio is very important feature for word wise script identification [21].

3 Pixels Ratio (PR): It is defined as the ratio between the on pixels of an input image after fill holes to its total number of pixels before fill holes, as illustrated in Fig. 5,. The value of the pixel ratio is a real number.

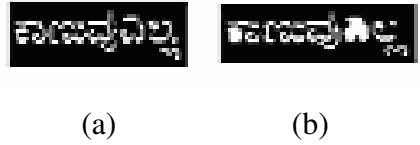

(a)          (b)

**FIGURE 5:** Shows Hole Fill Operation (a) Input Kannada Word, (b) After Hole Fill Operation of (a)

$$PixelsRatio = \frac{\sum onpixels(b)}{Size(a)}$$

4. Eccentricity: It is defined as the length of minor axis divided by the length of the major axis of a connected component of an image [20]. This is also a real valued function.

$$Average\_eccentricity = \frac{1}{N}\sum_{i}^{N} eccentricity(Component_i)$$

5. Extent: It is a real valued function; defined as the proportion of the pixels in the bounding box that are also in the region. It can be computed as area divided by the area of the bounding box.

$$Average\_extent = \frac{1}{N}\sum_{i}^{N} extent(Component_i)$$

The sample feature vectors of Devnagari, English numerals and Kannada scripts are as under.
     Devnagari = [0.3333   0.3333   0.3333   0.3333   0.3333   0.1846   0.9868   0.4872]
     English   = [0.1611     0      0.1384     0      0.2995   1.5515   0.7755   0.4030]
     Kannada  = [0.0172   0.1510   0.1682   0.1550   0.1805   0.7704    0.7184   0.4414]

**3.2. K-Nearest neighbour Classifier**
K-nearest neighbour is a supervised learning algorithm. It is based on minimum distance (Euclidian distance metric is used) from the query instance to the training samples to determine the k- nearest neighbours. After determining the k nearest neighbours, we take simple majority of these k-nearest neighbours to be the prediction of the query instance. The experiment is carried out by varying the number of neighbours (K= 3, 5, 7) and the performance of the algorithm is optimal when K = 3.

## 4. PEOPOSED ALGORITHM
The various steps involved in the proposed algorithm are as follows

1. Pre-process the input image i.e. binarisation using Otsu's method, and remove speckles using morphological opening.
2. Carry out the line wise and word wise segmentation based on horizontal and vertical projection profiles.
3. Carry out the morphological erosion and opening by reconstruction using the line structuring element in vertical, horizontal, left and right diagonal directions and perform the fill hole operation.
4. Compute the average pixel densities of the resulting images of step 3.
5. Compute the ratio of on pixels left after performing fill hole operation on input image to its size.
6. Compute the aspect ratio, eccentricity and extent of all the connected components of an input image and obtain their average values.
7. Classify the new word image based on the nearest neighbour and k-nearest neighbour classifiers.

## 5. RESULTS AND DISCUSSION

For experimentation, a sample image of size 256x256 pixels is selected manually from each document page and created a first data set of 2450 word images by segmentation. Out of these 2450 word images, Kannada, Devnagari are 750 each and Telugu and English numerals are 600 and 350 respectively. The second data set of 175 handwritten English numerals is used to test the potentiality of the proposed algorithm for script identification of handwritten numerals versus printed text words.

The classification accuracy achieved in identifying the scripts of first and second data set is presented in Table 2, 3, 4 and 5. Experimentally (based on principal component analysis), it is observed that, the right and left diagonal reconstruction features are not dominant and other six features are leading to retain the accuracy as reported. Although the primary aim of this paper is achieved, that is the word-wise script identification in bilingual documents; the fact is that, normally printed documents font sizes and styles are less varied. We therefore conducted a third set of experiment on 150 word images to test the sensitivity of the algorithm towards different font sizes and styles. These words are first created in different fonts using DTP packages, and then printed from a laser printer. The printed documents are scanned as mentioned earlier. On most commonly used five fonts of Kannada, Hindi and English are considered for experiment. For each font 10 word images are considered varying in font size from 10 to 36. Out of these 150 word images, Kannada, Devnagari and English numerals are 50 each. The Kannada font styles used are KN-TTKamanna, TTUma, TTNandini ,TTPadmini and TT-Pampa. The Devnagari font styles considered are DV-TTAakash, TTBhima, TTNatraj, TTRadhika, and TTsurekh. Times New Roman, Arial, Times New Roman italic, Arial Black and Bookman Old Style of English numerals are used for font and size sensitivity testing. It is noticed that, script identification accuracy achieved for third data set is consistent.

In the reported work of [4, 9, 12], it is mentioned that, the error rate is more when the word size is less than 3 characters. Our algorithm works for even single character words, but it fails when words like से ,हो, marks like "|" and broken sirorekha's are encountered in Devnagari. The touched, broken and bold face words of Kannada and Telugu are not recognized correctly because of loss in aspect ratio. Arial Black English numerals of size more than 16 points also misclassified. The sample test words and misclassified words are shown in Fig. 7 and 8.The proposed algorithm is implemented in MATLAB 6.1. The average time taken to recognize the script of a given word is 0.1547 seconds on a Pentium-IV with 128 MB RAM based machine running at 1.80 GHz. Since, there is no work reported for script identification of numerals at word level, to the best of our knowledge. However, the proposed method is compared with [4, 9, 12 and 14] as shown in Table 1.

## 6. CONCLUSION AND FUTURE WORK

In this paper, we investigated a tool of morphological opening by reconstruction of the image based on the strokes present in different directions and regional descriptors for script identification at word level. The simplicity of the algorithm is that, it works only on the basic morphological operations and shows its novelty for font and size independent script identification. The morphological reconstruction approach is efficiently used for extracting only those components or characters containing directional strokes and their densities are used for discriminating of the scripts. Furthermore, our method overcomes the word length constraint of [4, 9, 12] and works well even for single component words with minimum number of features. This work is first of its kind to the best of our knowledge. This algorithm can be generalized because of the visual appearance of every text is distinct and hence directional distribution of strokes must be distinct at an appropriate threshold. This work can also be extended to other Indian regional languages.

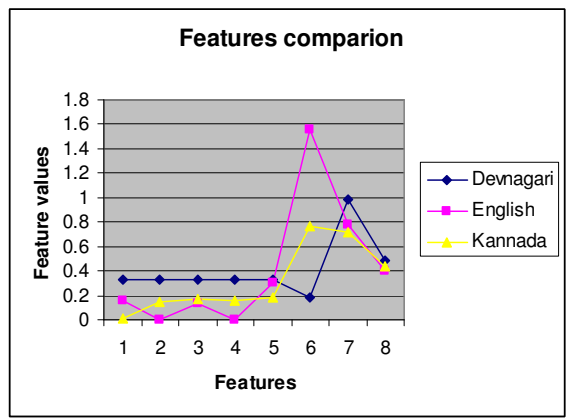

**FIGURE 6:** Shows Features Comparison of Three Scripts

| Methods Proposed by | Scripts | Accuracy in % | Time complexity | Comments |
|---|---|---|---|---|
| D.Dhanya | Tamil and Roman | 96.03 | Not reported | Algorithm suffers ,when the word length is less than four characters |
| U.Pal | Devnagari, Telugu and Roman | 96.72 | Not reported | Algorithm suffers ,when the word length is less than three characters |
| M.C.Padma | Kannada, Roman Devnagari | 95.66 | Not reported | Algorithm suffers ,when the word length is less than three characters |
| Peeta Basa Pati | Devnagari, Tamil Oriya | 97.33 | Not reported | Not reported |
| Proposed Method | Telugu and Roman | 97.60 | 0.15 seconds | Proposed algorithm works even for single character words. |
|  | Kannada and Roman | 96.68 |  |  |
|  | Devnagari and Roman | 98.58 |  |  |
|  | Telugu, Devnagari and Roman | 96.47 | 0.1547 seconds |  |
|  | Kannada, Devnagari and Roman | 95.54 |  |  |

**TABLE 1:** Comparative Study

| Script /language | NN | KNN |
|---|---|---|
| Telugu Vs. | 95.56% | 97.46 % |
| English numerals | 96.38% | 97.74% |
| Average | 95.97% | 97.6% |
| Kannada Vs. | 97.43% | 97.43 % |
| English numerals | 95.48% | 95.93% |
| Average | 96.45% | 96.68% |
| Devnagari Vs. | 98.53% | 98.53% |
| English numerals | 98.64% | 98.64% |
| Average | 98.58% | 98.58% |

**TABLE 2:** Script Recognition Results of Printed Telugu, Kannada and Devnagari Text Words with English Numerals

| Script | NN | KNN |
| --- | --- | --- |
| Telugu | 90.48% | 95.24% |
| Devnagari | 94.13% | 96.44% |
| English | 96.38% | 97.74% |

**TABLE 3:** Script Recognition Results of Telugu, Devnagari and Printed English Numerals

| Script | NN | KNN |
| --- | --- | --- |
| Kannada | 91.53% | 95.72% |
| Devnagari | 94.13% | 94.97% |
| English | 95.53% | 95.93% |

**TABLE 4:** Script Recognition Results of Kannada, Devnagari and Printed English Numerals**.**

| Script /language | KNN |
| --- | --- |
| Telugu Vs. | 91.43% |
| English numerals | 94.68% |
| Average | 93.05% |
| Kannada Vs. | 93.58% |
| English numerals | 95.74% |
| Average | 94.66% |
| Devnagari Vs. | 98.53% |
| English numerals | 98.94% |
| Average | 98.73% |

**TABLE 5:** Script Recognition Results of Printed Telugu, Kannada and Devnagari Text Words with Handwritten English Numerals.

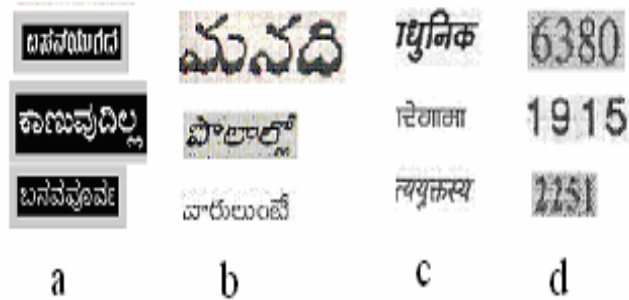

**FIGURE 7:** (a), (b), (c) and (d) are the Sample Test Images of Kannada, Telugu, Devnagari and English Numerals

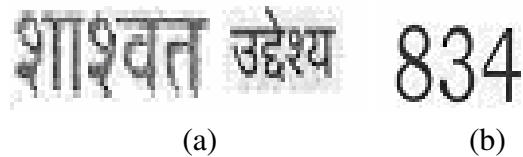

(a)                                    (b)

**FIGURE 8:** Misclassified Sample Word Images (a) Broken Sirorekha's of Devnagari Script, (b) English Numeral of Arial font of Size 24 Points

## 7. REFERENCES

1. B.B.Chaudhuri and U.Pal," *An OCR system to read two Indian language scripts: Bangla and Devnagari (Hindi)*", In Proceedings of 4[th] ICDAR, Uhn. 18-20 August, 1997

B.V.Dhandra and Mallikarjun Hangarge

2. B.B.Chaudhuri and U.Pal. "*A complete printed Bangla OCR*", Pattern Recognition vol.31, pp 531-549, 1998

3. Santanu Chaudhury, Gaurav Harit, Shekar Madnani, R.B.Shet," *Identification of scripts of Indian languages by Combining trainable classifiers*", In Proceedings of ICVGIP 2000, Dec-20-22, Bangalore, India.

4. D Dhanya, A.G Ramakrishnan and Peeta Basa pati, "*Script identification in printed bilingual documents*," Sadhana, vol. 27, part-1, pp. 73-82, 2002

5. J. Hochberg, P. Kelly, T Thomas and L Kerns, "*Automatic script identification from document images using cluster-based templates*," IEEE Transactions Pattern Analysis and Machine Intelligence, vol.19, pp.176-181, 1997

6. Judith Hochberg, Kevin Bowers, Michael Cannon and Patrick Keely, "*Script and language identification for hand-written document images,*" IJDAR-1999, vol.2, pp. 45-52

7. B.V.Dhandra, V.S.Malemath, Mallikarjun Hangarge, Ravindra Hegadi, "*Skew detection in Binary image documents based on Image Dilation and Region labeling Approach*", In Proceedings of ICPR 2006, V. No. II-3, pp. 954-957

8. Vincent, L.," *Morphological gray scale reconstruction in image analysis: Applications and efficient algorithms*," IEEE Trans. on Image processing, vol.2, no. 2, pp. 176-201, 1993

9. M.C.Padma and P. Nagabhushan," *Identification and separation of text words of Kannada Hindi and English languages through discriminating features*", In Proceedings of NCDAR-2003, pp- 252-260. 2003

10. G.S.Peake and Tan, "*Script and language identification from document images*", In Proceedings of Eighth British Mach. Vision Conf., vol.2, pp. 230-233, Sept-1997

11. U.Pal and B.B.Chaudhuri, "*Script line separation from Indian Multi-script documents,*" 5$^{th}$ ICDAR, pp.406-409, 1999

12. U.Pal. S.Sinha and B.B Chaudhuri, "*Word-wise Script identification from a document containing English, Devnagari and Telgu Text*," In Proceedings of NCDAR-2003, PP 213-220

13. S. Basavaraj, Patil and N.V.Subbareddy. "*Neural network based system for script identification in Indian documents*," Sadhana, vol. 27, part-1, pp. 83-97, 2002

14. Peeta Basa pati, S. Sabari Raju, Nishikanta Pati and A.G. Ramakrishnan, "*Gabor filters for document analysis in Indian Bilingual Documents*," In Proceedings of ICISIP-2004, pp. 123-126

15. P. Nagabhushan, S.A. Angadi and B.S. Anami," An Intelligent Pin code Script Identification Methodology Based on Texture Analysis using Modified Invariant Moments," In Proceedings of ICCR-2005, pp. 615-623

16. A.L.Spitz, "*Determination of the script and language content of document images,*" IEEE Transactions on Pattern Analysis and Machine Intelligence, Vol. 19, pp.234-245, 1997

17. A. L. Spitz, "*Multilingual document recognition Electronic publishing, Document Manipulations*, and Typography," R. Furuta ed. Cambridge Uni. Press, pp. 193-206, 1990

18. T.N.Tan, "*Rotation invariant texture features and their use in automatic script identification,*" IEEE Transactions on Pattern Analysis and Machine Intelligence, vol. 20, pp.751-756, 1998

19. S. Wood. X. Yao. K.Krishnamurthi and L.Dang    "*Language identification from for printed text independent of segmentation*," In Proceedings of International conference on Image Processing, pp. 428-431, 1995

20. Dengsheng Zhang, Guojun Lu, "*Review of shape representation and description techniques*," Pattern Recognition, vol. 37, pp. 1-19, 2004

21. Annop M. Namboodri, Anil K Jain, " *Online handwritten script identification*", IEEE Transaction on Pattern Analysis and Machine Intelligence, vol. 26,no.1,pp. 124-130, 2004

22. N. Otsu, " *A Threshold Selection Method from Gray-Level Histogram*" , IEEE Transaction Systems, Man, and Cybernetics, vol.9,no.1,pp.62-66,1979